\documentclass[twoside,11pt]{article}

\usepackage[arxiv-submission]{melba}

\usepackage{mwe} 

\usepackage{algorithm}
\usepackage{algorithmicx}

\usepackage[noend]{algpseudocode}
\usepackage[noend]{algpseudocode} 
\usepackage{graphicx}
\usepackage{epsfig}
\usepackage{amssymb}
\usepackage{mwe}
\usepackage{acro}
\usepackage{xcolor,colortbl}
\usepackage{tabularx}
\usepackage{relsize}
\usepackage{pifont}
\usepackage{booktabs} 
\usepackage{multirow}
\usepackage{multicol}
\usepackage{adjustbox}
\usepackage{float}
\usepackage{url}
\usepackage{algpseudocode}

\usepackage{color}

\makeatletter
\def\algbackskip{\hskip-\ALG@thistlm}
\makeatother

\usepackage[flushleft]{threeparttable}

\usepackage{amsmath,amsfonts}

\DeclareAcronym{ROI}{
short=ROI,
long=region of interest,
}

\DeclareAcronym{IOU}{
short=IOU,
long=intersection over union,
}

\DeclareAcronym{cIOU}{
short=cIOU,
long=circle intersection over union,
}

\DeclareAcronym{DoF}{
short=DoF,
long=degrees of freedom,
}

\DeclareAcronym{CPL}{
short=CPL,
long=Center Point Localization,
}

\melbaheading{2022:025}{https://www.melba-journal.org/papers/2022:025.html}{2022}{1-18}{02/2022}{08/2022}{Yao, Qu, Long, Liu, Deng, Tian, Xu, Jha, Asad, Bao, Zhao, Fogo, Landman, Yang, Chang and Huo}{}{}

\ShortHeadings{Compound Figure Separation of Biomedical Images}{Yao et al.}
\firstpageno{1}

\title{Compound Figure Separation of Biomedical Images: Mining Large Datasets for Self-supervised Learning}

\author{\name Tianyuan Yao \email tianyuan.yao@vanderbilt.edu \\  
	\addr Vanderbilt University, Department of Computer Science, Nashville, TN, USA 37215
	\AND
	\name Chang Qu \email chang.qu@vanderbilt.edu \\
	\addr Vanderbilt University, Department of Computer Science, Nashville, TN, USA 37215\AND
	\name Jun Long \email junlong@csu.edu.cn \\
	\addr Central South University, Big Data Institute, Changsha, Hunan, China 410083\AND
	\name Quan Liu \email quan.liu@vanderbilt.edu \\
	\addr Vanderbilt University, Department of Computer Science, Nashville, TN, USA 37215\AND
	\name Ruining Deng \email r.deng@vanderbilt.edu \\
	\addr Vanderbilt University, Department of Computer Science, Nashville, TN, USA 37215\AND
	\name Yuanhan Tian \email yuanhan.tian@vanderbilt.edu \\
	\addr Vanderbilt University, Department of Computer Science, Nashville, TN, USA 37215\AND
	\name Jiachen Xu \email jiachen.xu@vanderbilt.edu \\
	\addr Vanderbilt University, Department of Computer Science, Nashville, TN, USA 37215\AND
	\name Aadarsh Jha \email aadarsh.jha@vanderbilt.edu \\
	\addr Vanderbilt University, Department of Computer Science, Nashville, TN, USA 37215\AND
	\name Zuhayr Asad \email zuhayr.asad@vanderbilt.edu \\
	\addr Vanderbilt University, Department of Computer Science, Nashville, TN, USA 37215\AND
	\name Shunxing Bao \email shunxing.bao@vanderbilt.edu \\
	\addr Vanderbilt University, Department of Electrical and Computer Engineering, Nashville, TN, USA 37215\AND
	\name Mengyang Zhao \email mengyang.zhao@dartmouth.edu \\
	\addr Dartmouth College, Hanover, NH, USA 03755\AND
	\name Agnes B. Fogo \email agnes.fogo@vumc.org \\
	\addr Vanderbilt University Medical Center, Department of Pathology, Nashville, TN, USA 37215\AND
	\name Bennett A. Landman \email bennett.landman@vanderbilt.edu \\
	\addr Vanderbilt University, Department of Electrical and Computer Engineering, Nashville, TN, USA 37215\AND
	\name Haichun Yang \email haichun.yang@vumc.org \\
	\addr Vanderbilt University Medical Center, Department of Pathology, Nashville, TN, USA 37215\AND
	\name Catie Chang \email catie.chang@vanderbilt.edu \\
	\addr Vanderbilt University, Department of Electrical and Computer Engineering, Nashville, TN, USA 37215\AND
	\name Yuankai Huo \email yuankai.huo@vanderbilt.edu \\
	\addr Vanderbilt University, Department of Electrical and Computer Engineering, Nashville, TN, USA 37215
}

\begin{document}

\maketitle

\begin{abstract}
With the rapid development of self-supervised learning (e.g., contrastive learning), the importance of having large-scale images  (even without annotations) for training a more generalizable AI model has been widely recognized in medical image analysis. However, collecting large-scale task-specific unannotated data at scale can be challenging for individual labs. Existing online resources, such as digital books, publications, and search engines, provide a new resource for obtaining large-scale images. However, published images in healthcare (e.g., radiology and pathology) consist of a considerable amount of compound figures with subplots. In order to extract and separate compound figures into usable individual images for downstream learning, we propose a simple compound figure separation (SimCFS) framework without using the traditionally required detection bounding box annotations, with a new loss function and a hard case simulation. Our technical contribution is four-fold: (1) we introduce a simulation-based training framework that minimizes the need for resource extensive bounding box annotations; (2) we propose a new side loss that is optimized for compound figure separation; (3) we propose an intra-class image augmentation method to simulate hard cases; and (4) to the best of our knowledge, this is the first study that evaluates the efficacy of leveraging self-supervised learning with compound image separation. From the results, the proposed SimCFS achieved state-of-the-art performance on the ImageCLEF 2016 Compound Figure Separation Database. The pretrained self-supervised learning model using large-scale mined figures improved the accuracy of downstream image classification tasks with a contrastive learning algorithm.  
The source code of SimCFS is made publicly available at \url{https://github.com/hrlblab/ImageSeperation}. 
\end{abstract}


\section{Introduction}
\begin{figure}[h]
\begin{center}
\includegraphics[width=0.8\linewidth]{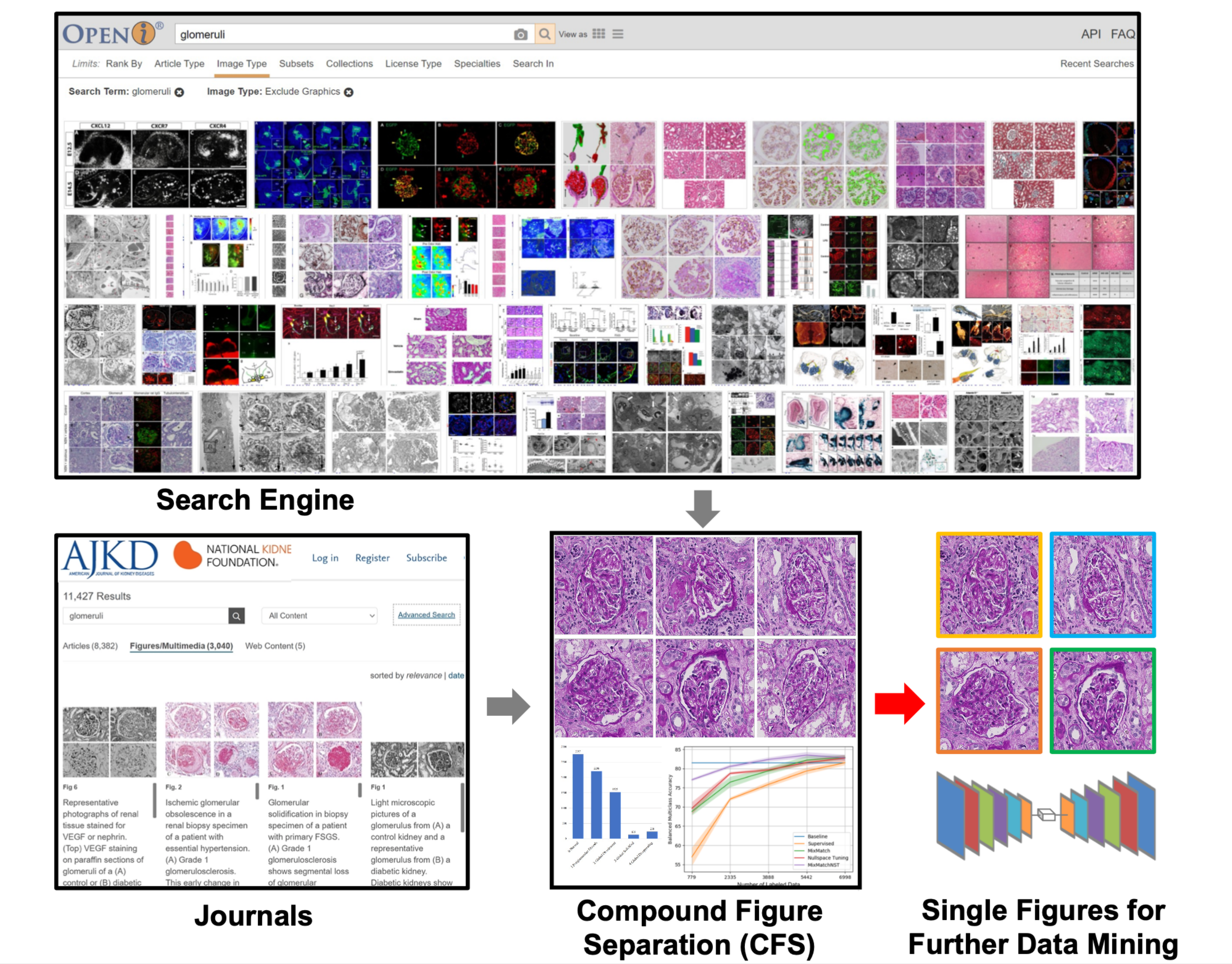}
\end{center}
   \caption{\textbf{Value of compound figure separation.} This figure shows the hurdle (red arrow) of training self-supervised machine learning algorithms directly using large-scale biomedical image data from biomedical image databases (e.g., NIH OpenI) and academic journals (e.g., AJKD). When searching desired tissues  (e.g., search ``glomeruli"), a large amount of data are compound figures. Such data would advance medical image research via recent self-supervised learning algorithms, such as self-supervised learning, contrasting learning, and auto encoder networks~\cite{huo2021ai}}
\label{fig:idea}
\end{figure}

Self-supervised learning algorithms (e.g., contrastive learning) allow deep learning models to learn effective image representations from large-scale unlabeled data~\citep{celebi2016unsupervised,sathya2013comparison,chen2020simple}. Thus, the important role of having large-scale images  (even without annotations) for training a more generalizable AI model has been widely recognized in medical image analysis. Even unannotated medical images can be difficult to obtain at scale for individual labs~\citep{zhang2017deep}. Fortunately, online resources (e.g., NIH  Open-i$^\circledR$~\citep{demner2012design} search engine, academic images released by journals) have provided a cost-effective and scalable way of obtaining large-scale images. However, the images from such resources consist of a considerably large amount of compound figures with subplots that cannot be directly used by modern self-supervised learning approaches (Fig~\ref{fig:idea}). To make the data useful, we need to extract individual subplots from the compound figure, with compound figure separation algorithms~\citep{lee2015dismantling}.

Recent contrastive learning methods have demonstrated advantages in pretraining a more generalizable deep learning model using large-scale unannotated individual images. However, the web-mined images from medical literature and search engines are not necessarily single images that can be directly used for contrastive learning. Therefore, the proposed SimCFS framework can be used to separate such compound images into individual images as unannotated training data for self-supervised learning. 

\begin{figure}[h]
\centering
\includegraphics[width=0.8\linewidth]{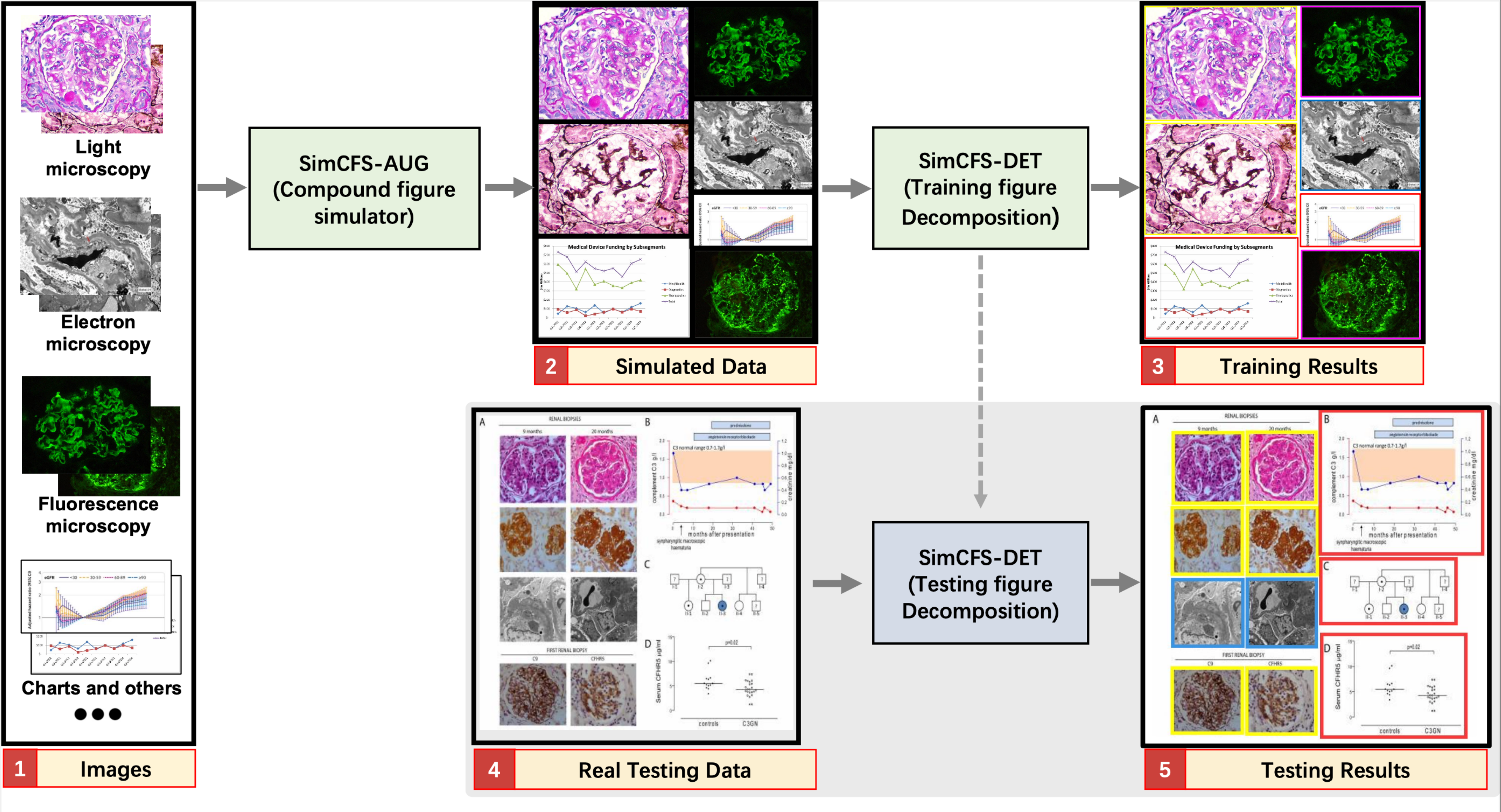}
\caption{\textbf{The overall workflow of the proposed simple compound figure separation (SimCFS) workflow.} In the training stage, SimCFS only requires individual images from different categories. The pseudo compound figures are generated from the proposed augmentation simulator (SimCFS-AUG). Then, a detection network (SimCFS-DET) is trained to perform compound figure separation. In the testing stage (the gray panel), only the trained SimCFS-DET is required for separating the images.}
\label{fig:network}
\end{figure}
Various compound figure separation approaches have been developed~\citep{davila2020chart,lee2015detecting,apostolova2013image,tsutsui2017data,shi2019layout,jiang2021two,huang2005associating}, especially with recent advances in deep learning. However, previous approaches typically required resource extensive bounding box annotation to form the problem as a training detection task. In this paper, we propose a simple compound figure separation (SimCFS) framework that minimizes the need for bounding box annotations in compound figure separation. Briefly, the contribution of this study is four-fold:

$\bullet$ We introduce a simulation-based training framework that minimizes the need of resource extensive bounding box annotations.

$\bullet$ We propose a new Side loss, which is an optimized detection loss for figure separation.

$\bullet$ We propose an intra-class image augmentation method to mimic the hard cases of compound images without clear boundaries.

$\bullet$ To the best of our knowledge, this is the first study that evaluates the efficacy of leveraging self-supervised learning with compound image separation.

We apply our technique to conduct compound figure separation for renal pathology (in-house data) as well as on the ImageCLEF 2016 Compound Figure Separation Database (publicly available). Glomerular phenotyping~\citep{koziell2002genotype} is a fundamental task for efficient diagnosis and quantitative evaluations in renal pathology. Recently, deep learning techniques have played increasingly important roles in renal pathology to reduce clinical working load of pathologists and enable large-scale population based research~\citep{gadermayr2017cnn,bueno2020glomerulosclerosis,govind2018glomerular,kannan2019segmentation,ginley2019computational}. Due to the lack of a publicly available dataset for renal pathology, it is appealing to extract large-scale glomerular images from public databases (e.g., NIH Open-i$^\circledR$ search engine) for downstream self-supervised or semi-supervised learning~\citep{huo2021ai}. Meanwhile, the Image-CLEF 2016 dataset consists of various types of organs, and resources of large-scale medical images, which is arguably the most widely used testbed for compound image separation tasks. Both cohorts are used to evaluate the performance of different methods.

This work is extended from our conference paper~\citep{yao2021compound} with the new efforts listed below: (1) we included more technical and evaluation details for the proposed method; (2) More comprehensive literature review and related work have been introduced; (3) We performed more rigorous evaluation (five-fold cross-validation) during the evaluation stages; (4) We conducted more comprehensive evaluation with more baseline compound image generation and separation methods  (e.g., \cite{tsutsui2017data}); (5) We evaluated the efficacy of leveraging self-supervised learning with compound image separation by evaluating with both supervised and semi-supervised methods; (6) Our web mined glomerular dataset (20,000 images), as well as the source code of SimCFS, are released to public in the paper.

\section{Related Work}
\subsection{Compound Figure Separation}
In biomedical articles, about 40-60$\%$ of figures are multi-panel~\citep{kalpathy2015evaluating}. Several methods have been proposed in the document analysis community that envolve, extracting figures and their  semantic information. For example, \cite{huang2005associating} presented their recognition results of textual and graphical information in literary figures. \cite{davila2020chart} presented a survey of approaches of several data mining pipelines for future research.

\subsubsection{Traditional vision approaches}
In order to collect scientific data massively and automatically, various approaches have been proposed in the prior arts\citep{10.1093/bioinformatics/btx611,10.1007/978-3-319-65813-1_20,lee2015dismantling}. For example, \cite{lee2015detecting} proposed an SVM-based binary classifier to distinguish completed charts from visual markers, such as labels, legend, and ticks. \cite{apostolova2013image} proposed a figure separation method via a capital index. These traditional computer vision approaches were commonly performed on the figure's grid-based layout. Thus, the separation was usually accomplished by simple horizontal and vertical cuts based on the image boundary information.

\subsubsection{Deep learning Methods}
In the past few years, deep learning based algorithms, especially convolutional neural networks (CNNs), have provided considerably superior performance in extracting and separating subplots from from compound images. \cite{tsutsui2017data} proposed a CNN based approach that treated compound figure segmentation as an object localization problem by estimating the bounding boxes of subplots. This was one of the earliest deep learning-based approaches to achieve compound figure separation via a deep convolutional neural network. Tsutsui et al. applied the You Only Look Once (YOLO) Version 2, a CNN based detection network, which utilized a single convolutional network to predict bounding boxes and class probabilities simultaneously. They also implemented training on artificially constructed datasets and reported superior performances on ImageCLEF dataset~\citep{GSB2016}. \cite{shi2019layout} developed a multi-branch output CNN to predict the irregular panel layouts and provided augmented data to drive learning. Their network separated compound figures of different sizes of structures with better accuracy. 

More recently, anchor-based approaches have attracted great attentions in the object detection field due to their concise network architectures and high computational efficiency. The introducing of anchor has prior knowledge to object distribution which is also closer to the compound figure situation. YOLOv4 was used by \cite{jiang2021two} to achieve a superior detection performance.  They combined a traditional vision method with high performance of deep learning networks by detecting the sub-figure label and then optimizing the feature selection process in the sub-figure detection. Now, YOLO has been updated to V5, which inherited the advantages of YOLOv4~\citep{bochkovskiy2020yolov4}. YOLOv5 integrated spatial pyramid pooling with new data enhancement methods like Mosaic training, balanced model size and detection speed which achieved faster detection speed and higher accuracy.


\subsection{Self-supervised learning method}
Supervised learning refers the usage of a set of input variables to predict the value of a labeled output variable. It requires labeled data (like an answer key that the model can use to evaluate its performance). Conversely, self-supervised learning~\citep{celebi2016unsupervised} refers to inferring underlying patterns from an unlabeled dataset without any reference to labeled outcomes or predictions.

Recently, a new family of self-supervised representation learning, called contrastive learning, shows its superior performance in various vision tasks ~\citep{wu2018unsupervised,noroozi2016unsupervised,zhuang2019local,hjelm2018learning}. Learning from large-scale unlabeled data, contrastive learning can learn discriminative features for downstream tasks. SimCLR~\citep{chen2020simple} maximizes the similarity between images in the same category and repels the representations of different category images. \cite{wu2018unsupervised} uses an offline dictionary to store all data representation and randomly selects training data to maximize negative pairs. MoCo~\citep{he2020momentum} introduces a momentum design to maintain a negative sample pool instead of an offline dictionary. Such works demand a large batch size in order to include sufficient negative samples. To eliminate the needs of negative samples, BYOL ~\citep{grill2020bootstrap} was proposed to train a model with an asynchronous momentum encoder. Recently, SimSiam~\citep{chen2020exploring} was proposed to further eliminate the momentum encoder in BYOL, allowing  for less GPU memory consumption.


\section{Methods}
The overall framework of SimCFS is presented in Fig. ~\ref{fig:network}. The training stage of SimCFS contains two major steps: (1) compound figure simulation, and (2) sub-figure detection. In the training stage, the SimCFS network can be trained with either a binary (background and sub-figure) or multi-class setting. The purpose of the compound figure simulation is to achieve collecting large-scale training compound images in an annotation free manner. In the testing stage, only the detection network is needed, where the output will be the bounding boxes of the sub-figures which shall enable us to crop those images in a fully automatic manner. The binary setting detector can serve as a compound figure separator, while the multi-class detector can be used for web image mining for images of concerned categories.

\subsection{Anchor-based detection}
YOLOv5, the latest version in the YOLO family~\citep{bochkovskiy2020yolov4}, is employed as the backbone network for sub-figure detection. The rationale for choosing YOLOv5 is that the sub-figures in compound figures are typically located in horizontal or vertical orders. Herein, the grid-based design with anchor boxes is well adaptable to our application. A new Side loss is introduced to the detection network that further optimizes the performance of compound figure separation. 

\subsection{Compound figure simulation}
Our goal is to only utilize individual images, which are non-compound images with weak classification labels in training a compound image separation method. In previous studies, the same task typically requires stronger bounding box annotations of subplots using real compound figures. In compound figure separation tasks, a unique advantage is that the sub-figures are not overlapped. Moreover, their spatial distributions are more ordered as compared with natural images in object detection. Therefore, we propose to directly simulate compound figures from individual images as the training data for the downstream sub-figure detection.

\begin{figure*}[t]
\begin{center}

\includegraphics[width=0.9\linewidth]{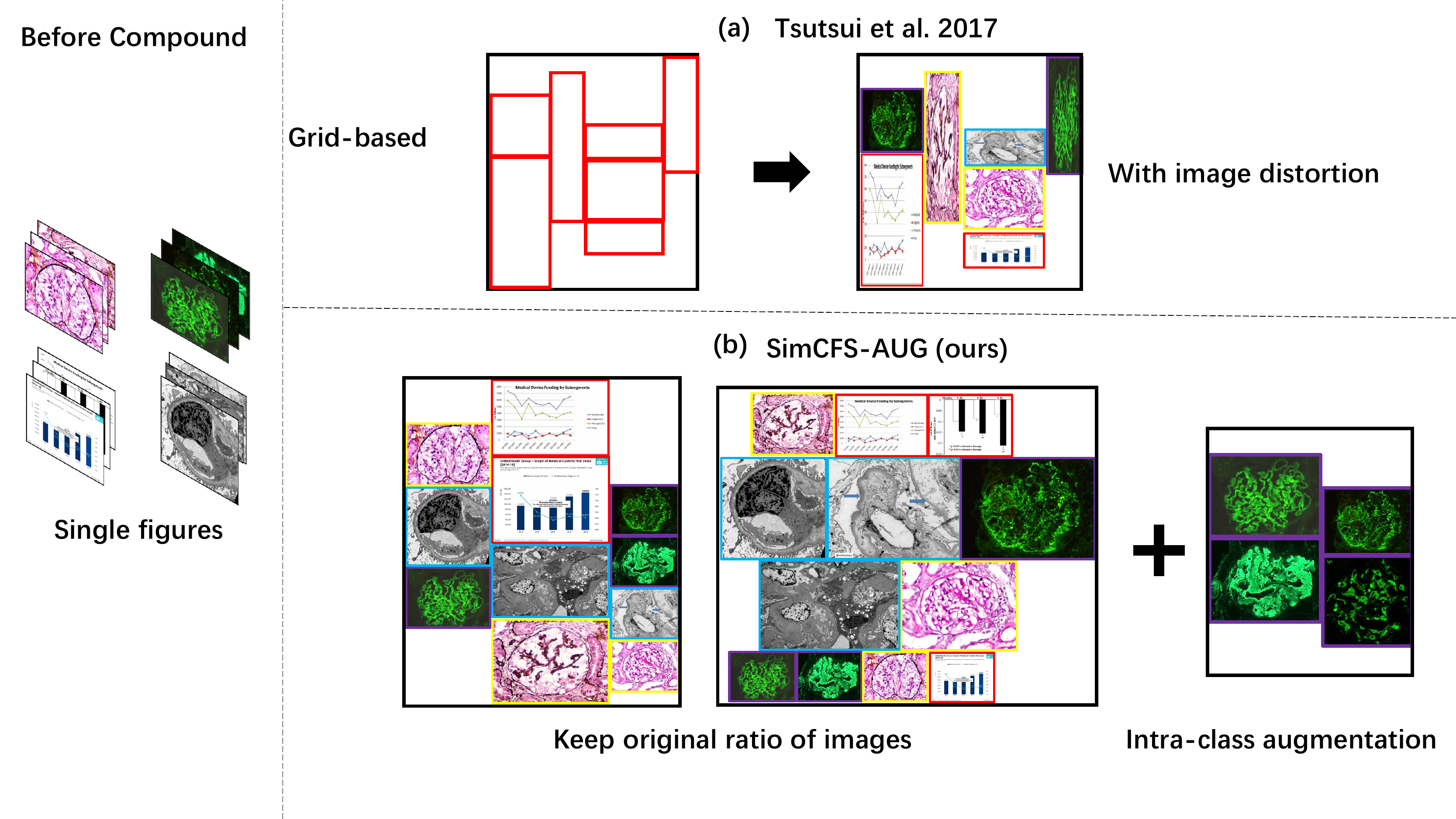}
\end{center}
   \caption{\textbf{Compound figure simulation.} 
   (a) The upper panel shows the previously proposed compound figure synthesis strategy. It first generates the figure grids and then fills with images that have undergone image distortion, which is unusual in real compound figures. (b) The lower panel presents the proposed SimCFS-AUG compound figure simulator. It keeps the original ratio of individual images in an adaptive manner. Beyond this step of keeping original ratios, an intra-class augmentation is introduced to simulate the hard cases in which the boundaries are not explicitly visible between similar subplots. (Bounding boxes are displayed for visualization and are not actually visible in the training data)}
\label{fig:method}
\end{figure*}

\cite{tsutsui2017data} proposed a compound figure synthesis approach (Fig.~\ref{fig:method}). The method first randomly samples a number of rows and generates random heights for each row. Then a random number of single figures fills the empty template. However, the single figures are naively resized to fit the template, with large distortion (Fig.~\ref{fig:method}).

\begin{algorithm}
\caption{Compound figure simulation}
     \textbf{Input:} \\
 \hspace*{\algorithmicindent}   Single images $X_{i}$ in $k$ classes\\
 \hspace*{\algorithmicindent}Set of training input indices with known labels $L_{1}, L_{2}, ..., L_{k}$ \\
    \textbf{Output:} \\
 \hspace*{\algorithmicindent}Compound figure $\overline{C}_{j}$\\
 \hspace*{\algorithmicindent}Annotation file $A_{j}$
\begin{algorithmic}[1]
\For{each pseudo compound figure $\overline{C}_{j}$}
    \State \textbf{Stage 1:} Space initialize \Comment{Multi real world case simulation}
    \State $\textit{Layout} \gets \textit{row-restricted or column-restricted}$
    \State $\textit{Classes} \gets \textit{multi or intra}$
    \Comment{Add intra-class augmentation}
    \State $\textit{Number of rows/columns} \gets \textit{$n \in [2,5]$ }$
    \If{layout is row-restricted} \Comment{Keep close to real world aspect ratio}
        \State $ \textit{Width} W_{\overline{C}_{j}} \gets 640,  \textit{Height} H_{\overline{C}_{j}}  \gets  \sum_{p=1}^n H_{p} \quad \textit{while} \quad  \frac{3}{4} \leq aspect\ ratio \leq \frac{4}{3} $
    \State \Comment{Each row's height $H_{1},...H_{p}$ should be in certain range}
    \ElsIf{layout is column-restricted}
        \State $ \textit {Height}H_{\overline{C}_{j}} \gets 640, \textit {Width}W_{\overline{C}_{j}}  \gets   \sum_{q=1}^n W_{q} \quad \textit{while} \quad  \frac{3}{4} \leq aspect\ ratio \leq \frac{4}{3}$
    \State \Comment{Each column's width $W_{1},...W_{q}$ should be in certain range}
    \EndIf
    \State \textbf{Stage 2:} Fit in preset space
    \For{row/column in  n}
        \If{Classes is multi}
            \State Create ImagePool I, for images $X_{i}$ in I, i $\in L_{1}, L_{2}, ..., L_{k}$
        \ElsIf{Classes is intra}
            \State Create ImagePool I, for images $X_{i}$ in I, i $\in L_{m}, m \in [1,k]$
        \EndIf
        \State Random fill in resized images from ImagePool (keeping original ratio) 
        \State Save resized $w_{i}', h_{i}'$, center position $x_{i}, y_{i}$ to $A_{j}$
    \EndFor
    \State \textbf{Stage 3:} Output: compound figure $\overline{C}_{j}$, annotation $A_{j}$
\EndFor
\end{algorithmic}
\end{algorithm}

Inspired by prior arts~\citep{tsutsui2017data}, we propose a simple augmentation strategy that is specific to compound figure separation data, called SimCFS-AUG, to perform compound figure simulation. The inputs of the simulator are single images with specified classes. Two groups are generated when simulating compound figures; these groups are row-restricted and column-restricted. The length of each row or column is randomly generated within a certain range. Then, images from our database are randomly selected and concatenated together to fit in the preset space. As opposed to previous studies, the original ratio of individual images is kept within our SimCFS-AUG simulator so as to mimic more realistic common compound images without distortion in individual images.

\begin{figure*}[t]
\begin{center}
\includegraphics[width=0.8\linewidth]{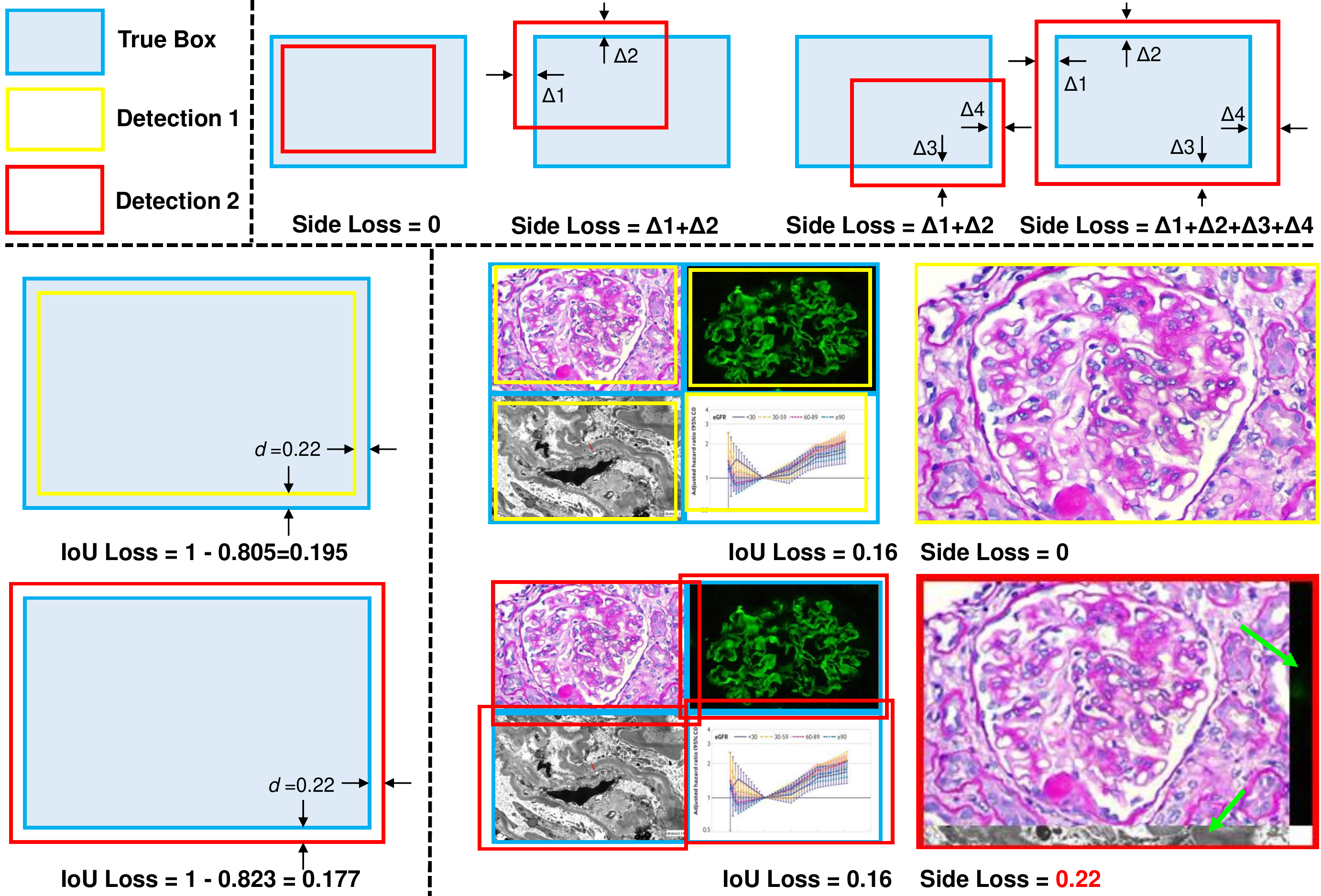}
\end{center}
\caption{\textbf{Proposed Side loss for figure separation}. The upper panel shows the principle of side loss, in which penalties only apply when vertices of detected bounding boxes are outside of true box regions. The lower left panel shows the bias of current IoU loss towards over detection. When an under detection case (yellow box) and an over detection case (red box) have the same margins ($d$), from predicted to true boxes, the over detection has the smaller loss (larger IoU). The lower right panel shows the under detection and over detection examples of the compound figure separation, with the same IoU loss. Side loss is proposed to break IoU loss, given the results in the yellow boxes are less contaminated by nearby figures than the results in the red boxes (green arrows).}
\label{fig:sideloss}
\end{figure*}

\subsection{Side loss for compound figure separation}
For object detection on natural images, there is no specific preference between over detection and under detection as objects can be randomly located and even overlapped. In medical compound images, however,objects are typically closely attached to each other without overlapping. In this case, over detection would introduce undesired pixels from the nearby plots (Fig. ~\ref{fig:sideloss}), which are not ideal for downstream deep learning tasks. Unfortunately, over detection is often encouraged by the current Intersection Over Union (IoU) loss in object detection (Fig. ~\ref{fig:sideloss}), as compared with under detection.

In the SimCFS-DET network, we introduce a simple side loss, which will penalize over detection. We define a predicted bounding box as $B^p$ and a ground truth box as $B^g$, with coordinates: $B^p = (x^p_1,y^p_1,x^p_2,y^p_2) $,$\quad B^g = (x^g_1,y^g_1,x^g_2,y^g_2)$. The over detection penalty of vertices for each box is computed as:
\begin{equation}
\begin{aligned}
x^{\mathcal{I}}_1 = \max(0,x^g_1-x^p_1),
y^{\mathcal{I}}_1 = \max(0,y^g_1-y^p_1)\\
x^{\mathcal{I}}_2 = \max(0,x^p_2-x^g_2),
y^{\mathcal{I}}_2 = \max(0,y^p_2-y^g_2)
\end{aligned}
\end{equation}

Then, the Side loss is defined as:
\begin{equation}
\mathcal{L}_{side} = x^{\mathcal{I}}_1+y^{\mathcal{I}}_1+x^{\mathcal{I}}_2+y^{\mathcal{I}}_2
\end{equation}

The side loss is combined with canonical loss functions in YOLOv5, including bounding box loss (${L}_{box}$), object probability loss (${L}_{obj}$), and classification loss (${L}_{cls}$).\\ 
{$ \mathcal{L}_{total} = \lambda_1{L}_{box} + \lambda_2{L}_{obj} + \lambda_3{L}_{cls} + \lambda_4{L}_{side}
$}   ,where $\lambda_1$, $\lambda_2$, $\lambda_3$, $\lambda_4$ are constant weights to balance the four loss functions. Following YOLOv5's implementation \footnote{https://github.com/ultralytics/yolov5}, the parameters were set as $\lambda_1$ = ${box}\times(3/{nl})$, $\lambda_2$ = ${obj}\times{({imgsize}/640)^{2}\times(3/{nl})}$, $\lambda_3$ = $({cls}\times{num\_cls}/80)\times(3/{nl})$, where ${num\_cls}$ was the number of classes, ${nl}$ was the number of layers, and ${imgsize}$ was the image size.The $\lambda_4$ of the Side loss was empirically set to $\lambda_1/30$ across all experiments as the Side loss and Box loss are all based on the coordinates.

\begin{figure*}[h]
\centering
\begin{center}
\includegraphics[width=0.81\linewidth]{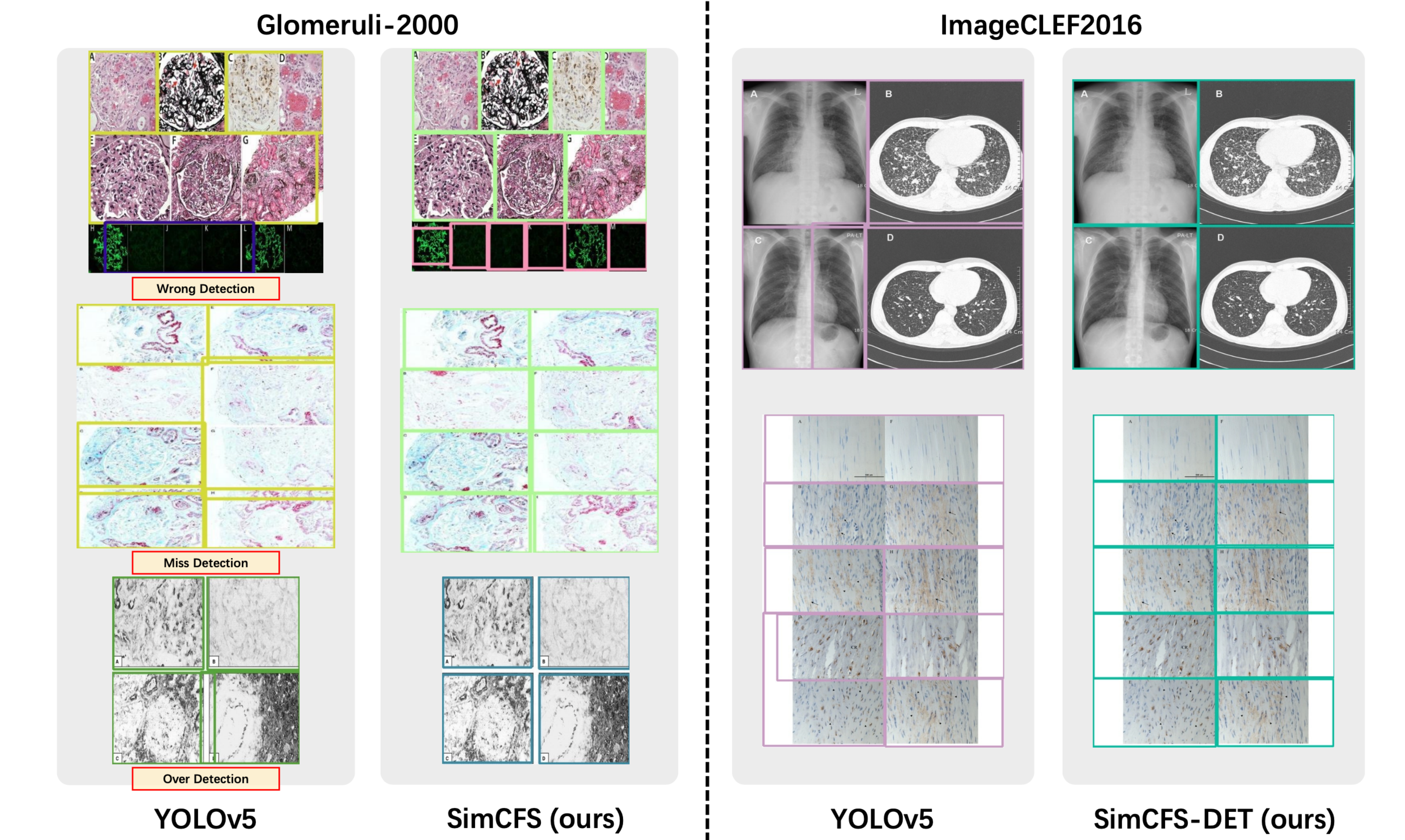}
\end{center}
   \caption{\textbf{Qualitative Results.} This figure shows the qualitative results of comparing proposed SimCFS approach with the YOLOv5 benchmark.}
\label{fig:results}
\end{figure*}

\section{Experimental Design}
\subsection{Data}
We collected two in-house datasets for evaluating the performance of different compound figure separation strategies. One compound figure dataset (called Glomeruli-2000) consisted of 917 training and 917 testing real figure plots from the American Journal of Kidney Diseases (AJKD), with keywords ``glomerular OR glomeruli OR glomerulus". Each figure was annotated manually with four classes, including glomeruli from (1) light microscopy, (2) fluorescence microscopy, (3) electron microscopy, and (4) charts/plots. 

To obtain individual images to simulate compound figures, we downloaded 5,663 single individual images from online resources. Briefly, we obtained 1,037 images from Twitter, and obtained 4,626 images from Google search, with five classes, including individual images from (1) glomeruli with light microscopy, (2) glomeruli with fluorescence microscopy, (3) glomeruli with electron microscopy, (4) charts/plots, and (5) others. The individual images were combined using the SimCFS-AUG simulator in order to generate 7,000 pseudo training images. 2,000 of the pseudo images (with multiple sub-figures) were simulated using intra-class augmentation. In addition, 2,947 individual images were further employed as training data. The implementation of SimCFS-DET was based on YOLOv5 with PyTorch implementations. Google Colab was used to perform all experiments in this study.

\subsection{Implement Details}
In the experiment setting, the parameters are empirically chosen. We set the learning rate to 0.01, weight decay to 0.0005 and momentum to 0.937. The input image size was set to 640, ${box}$ to 0.5, ${obj}$ to 1, ${cls}$ to 0.5, and the number of layers to 3. For our in-house datasets, we trained 50 epochs using a batch size of 64. For the imageCLEF2016 dataset~\citep{GSB2016}, we trained 50 epochs using a smaller batch size of 8.

\subsection{Evaluation Metrics}
Mean average precision was the primary metric used to evaluate detection performance. For a given threshold IOU, average precision was obtained by calculating the area under the 101-point interpolated precision-recall curve. Then, mean average precision ($AP$) is the mean of the average precision for IOU thresholds from 0.5 to 0.95 with a step size of 0.05. $AP_{50}$ is the average precision with an IOU threshold at 0.5. $AP_{75}$ is the average precision with an IOU threshold at 0.75. $AP_S$ is the mean average precision for small objects (area less than $32^2$). $AP_M$ is the mean average precision for medium objects (area between $32^2$ and $96^2$). Since no objects contained an area greater than $96^2$, the large mean average precision ($AP_L$) was not utilized.

\section{Results}
\subsection{Ablation Study}

In this ablation study, we evaluate the image separation performance via 917 real compound images with manual box annotations as testing data in \ref{table:ablation} and Fig.~\ref{fig:results}. For training, we assessed the performance of using 917 real compound training images (“Real Training Images”), as well as the performance when only using simulated training images (``Simulated Training Images''). 

From the result, the proposed Side loss consistently improves the detection performance by a decent margin. The proposed compound image simulation method (with intra-class self-augmentation) achieves superior performance as compared to the benchmarks.

\begin{table*}[t]
\caption{The ablation study with different types of training data.}
\begin{tabular}{p{3.5cm}<{\centering}|p{1.2cm}<{\centering}p{0.7cm}<{\centering}p{0.7cm}<{\centering}p{0.9cm}<{\centering}p{0.9cm}<{\centering}p{0.9cm}<{\centering}p{0.9cm}<{\centering}p{0.9cm}}
\toprule
Method & Training Data &SL & AUG & All  & Light & Fluo.  & Elec. &Chart \\
\midrule
YOLOv5 &$R$  &   &       & 69.8      & 77.1 & 71.3       & 73.4 & 57.4\\
SimCFS-DET (ours) &$R$   & \checkmark &      & \underline{79.2}  & \underline{86.1}      & \textbf{80.9}     & 84.2 & \textbf{65.8}  \\
\midrule
YOLOv5  &$\bar{S}$ &  &  & 63.8 & 76.4 & 60.1 & 72.5  & 46.8 \\
YOLOv5  &$S$ &  &  & 66.4  & 79.3 & 62.1  & 76.1 & 48.0\\
YOLOv5  &$S$ &  &\checkmark & 71.4 & 82.8 & 72.1  & 75.3  &47.1   \\
SimCFS (ours)  &$\bar{S}$ &\checkmark  &   & 68.9  & 77.1  & 66.8  & 82.5  & 49.1 \\
SimCFS (ours) &$S$   & \checkmark &   &69.4  & 77.6 & 67.1 & \underline{84.1}  & 48.8 \\
SimCFS (ours) &$S$  & \checkmark  & \checkmark     & \textbf{80.3}      & \textbf{89.9} & \underline{78.7}       & \textbf{87.4}  &\underline{58.8}\\
\bottomrule

\end{tabular}

\begin{tablenotes}
  \small
  \item *The best and second best performances are denoted by \textbf{bold} and \underline{underline}.
   \item *For training data, $R$ is using real compound figure while $S$ is using simulated images, $\bar{S}$ is using \cite{tsutsui2017data} grid-based synthetic method.
   \item*SL is the side loss, AUG is the intra-class self-augmentation.
   \item*ALL is the Overall mAP$_{0.5:.95}$, which is reported for all concerned classes, (light, fluorescence, \\ and electron microscopy).
\end{tablenotes}

\label{table:ablation}
\end{table*}

\begin{table}[t]

\caption{The results on ImageCLEF2016 dataset.}
\centering
\begin{tabular}{p{6cm}p{1.6cm}p{1.0cm}p{1.0cm}p{2.0cm}p{3.0cm}}
\toprule
Method      & Backbone          & mAP$_{0.5}$ & mAP$_{0.5:.95}$ \\
\midrule
\cite{tsutsui2017data}   & YOLOv2      & 69.8 & -       \\
\cite{tsutsui2017data}  & Transfer  & 77.3 & -       \\
\cite{zou2020unified}   & ResNet152      & 78.4 & -       \\
\cite{zou2020unified}  & VGG19          & 81.1 & -       \\
YOLOv5~\citep{bochkovskiy2020yolov4}  & YOLOv5   & 85.3 & 69.5   \\
SimCFS-DET (ours)   &YOLOv5  & 88.9 & 71.2  \\
SimCFS-DET esemble (ours) &YOLOv5  & \textbf{90.3} & \textbf{71.5}  \\
\bottomrule
\end{tabular}
\label{table:stateoftheart}
\end{table}

\subsection{Comparison with State-of-the-art}
We also compare CFS-DET with the state-of-the-art approaches including \cite{tsutsui2017data} and \cite{zou2020unified} using the ImageCLEF2016 dataset~\citep{GSB2016}. ImageCLEF2016 is the commonly accepted benchmark for compound figure separation, including total 8,397 annotated multi-panel figures (6,783 figures for training and 1,614 figures for testing). Table \ref{table:stateoftheart} shows the results of the ImageCLEF2016 dataset. The proposed CFS-DET approach consistently outperforms other methods by considering  evaluation metrics. Additionally, we applied five-fold cross validation to our model training using weighted boxes fusion as proposed by \citep{solovyev2021weighted}. To merge the bounding boxes results from the five predictions, the proposed method used the confidence scores of all of the proposed bounding boxes in order to construct the average boxes. Eventually, when combining SimCFS with the weighted boxes fusion (SimCFS-DET ensemble), the performance was further improved.

\subsection{Application on Contrastive Learning}
We demonstrate the application of our SimCFS framework and how it helps to provide massive biomedical image data and benefits further data analysis with self-supervised representation learning.

In this study, self-supervised contrastive learning was employed as an example downstream task for our SimCFS compound image separation approach. We demonstrate how our approach helps to provide massive biomedical image data and benefits further data analysis with self-supervised representation learning. To evaluate the performance of introducing separated images, a semi-supervised method was evaluated beyond the supervised benchmark to present the performance of using the same set of unannotated images as the contrastive learning approach.(Table \ref{table:performance}) Specifically, the stain and imaging modality classification task is employed to evaluate the performance of different approaches.

\subsubsection{Data}

We first collected 10,000 compound figures with the keywords `glomerular OR glomeruli OR glomerulus'. Then we used our SimCFS network to process all compound images to get more than 20,000 glomeruli pathologies obtained by different microscopy or in different stains with a confidence threshold of 0.7.

Other in-house data are 3,000 manually annotated glomeruli pathologies with seven classes, including glomeruli from (1) electron microscopy, (2) fluorescence microscopy,  and light microscopy with different stains of (3) PAS, (4) silver, (5) H\&E, (6) Masson and (7) other.

\subsubsection{Approach}
We used the SimSiam network~\citep{chen2020simple} as the baseline method of contrastive learning. 20,000 glomeruli pathologies were used to train the SimSiam network. Two random augmentations from the same image were used as training data. In all of our self-supervised pre-training, images for model training were resized to $224\times224$ pixels. We used the momentum SGD as the optimizer. The weight decay was set to 0.0005. The base learning rate was $lr=0.05$ and the batch size equals 64. The learning rate was $lr\times$BatchSize$/256$, which followed a cosine decay schedule~\citep{loshchilov2017sgdr}. 

To apply the self-supervised pre-training networks, we froze the pretrained ResNet-50 model by adding one extra linear layer which followed the global average pooling layer. When finetuning with the 3,000 manually annotated glomeruli data, only the extra linear layer was trained.To prevent model over-fitting, we applied 5-fold cross validation by dividing our data into 5 folders, using four of the five folders as training data and the other folder as validation. We used the SGD optimizer to train linear classifier with a based (initial) learning rate $lr$=30, weight decay=0, momentum=0.9, and batch size=64 (follows~\cite{chen2020exploring}). We trained linear classifiers for 100 epochs and selected the best model based on the validation set.

\subsubsection{Results}
Fine-tuning our pretrained SimSiam (Backbone:ResNet-50) on 2.3K labeled images is significantly better then training from scratch. Interestingly, our model also outperformed ResNet-50 models pretrained on ImageNet. Table \ref{table:performance} shows the results.

\begin{table}
\caption{Classification performance.}
\centering
\begin{tabular}{p{5cm}<{\centering}p{1.5cm}p{1.5cm}p{1.5cm}p{1.5cm}}
\toprule
Methods & Unlabeled Images & labeled Images  & F1 \quad \quad Score &Balanced Acc \\
\midrule
\multicolumn{5}{l}{\textbf{Supervised method:}}\\
Random Int &-  &2.3k  & 0.845 & 0.843\\ ImageNet Int  &-  &2.3k  & 0.888 & 0.883  \\ \midrule
\multicolumn{5}{l}{\textbf{Semi-supervised method:}}\\
Temporal Ensembling  & 20k  &2.3k & 0.892 & 0.885 \\
\midrule
\multicolumn{5}{l}{\textbf{Self-supervised method:}}\\
Simsiam  & -  &2.3k & 0.891 & 0.893 \\ 
Simsiam w.SimCFS &  20k  &2.3k & \textbf{0.900} & \textbf{0.904} \\ 
\bottomrule
\end{tabular}
\label{table:performance}
\begin{tablenotes}
  \small
   \item *For the supervised method, we trained the entire ResNet-50 (random initialized and ImageNet pretrained) from scratch with fully supervised learning.
\end{tablenotes}

\end{table}

\section{Discussion}
In this study, we develop a new compound image separation framework with the ultimate goal to advance downstream machine learning tasks. The recent contrastive learning methods demonstrated their advantages of pretraining a more generalizable deep learning model using large-scale unannotated individual images. However, the web-mined images from medical literatures and search engines are not necessarily single images that can be directly used for contrastive learning. Therefore, the proposed SimCFS can be used to separate such compound images into individual images as unannotated training data for self-supervised learning.

The YOLO method was employed since it was a broadly used anchor-based backbone in previous compound image separation algorithms. However, our framework is an open framework, where the YOLO method can be replaced by other object detection backbones (e.g., anchor-free methods) and even with an even better performance.

 The new application, through the optimization of both Side loss function and hard case simulation, proposes to improve the accuracy of image separation. Our proposed Side loss is designed based on the knowledge that there is no overlapping case in compound figures. By adding a penalty for the overestimated bounding box, the predictions are less overlapped as compared to the true box regions.


Secondly, with our compound figure simulation method, SimCFS can be trained with only synthetic compound figures which are generated by only a small quantity of annotated individual images. At the beginning of our experiment, when we synthesized row-restricted and column-restricted compound figures using images from all classes, the results were not as good as the real compound image data. To overcome such issues, we proposed the intra-class image augmentation method. By simulating those hard cases and adding the new intra-class compound figures to our previous synthesized data, the performance of the simulated training data has outperformed the real data by its large quantity and various simulated cases.

Recent advances in computer vision are due, to a large extent, to the growing size of annotated training data. However, one key limitation to the SimCFS network is that the ImageCLEF Medical dataset , the largest available dataset for compound figure separation, has only 7,000 images for training, which is much smaller than most modern object detection datasets. An important goal for this community could be to build up a much larger size dataset with multi-classes annotations like MRI, pathology, and charts etc. In this study, we assessed the promising application of SimCFS, which is to create large-scale unlabeled images for downstream contrastive learning. Using NIH OpenI, tens of thousands of free biomedical data can be achieved by searching the desired tissue types. The self-supervised learning strategy achieved better accuracy than the fully supervised approach with ImageNet initialization.

Several potential improvements for our SimCFS framework are as follows. First, we could further introduce image synthesis approaches to the proposed pipeline to obtain more unique images\. Furthermore, we can perform textual contents extractions for captions, notes and labels while separating figures. These data in multi-forms could benefit further data mining research.

\section{Conclusion}
In this paper, we introduced the SimCFS framework to extract images of interests from large-scale compounded figures with weak classification labels. The pseudo training data were built using the proposed SimCFS-AUG simulator. The anchor-based SimCFS-DET detection achieved state-of-the-art performance by introducing a simple side loss. Additionally, our SimCFS framework provided cost-efficient and large-scale unannotated images to train un-/self-supervised representation learning methods (e.g., SimSiam). It achieved better performance than ImageNet's supervised pre-trained counterparts in classification tasks.

\acks{This work was supported in part by NIH NIDDK DK56942(ABF) and NSF CAREER 1452485 (Landman).}

%
\ethics{The work follows appropriate ethical standards in conducting research and writing the manuscript, following all applicable laws and regulations regarding treatment of animals or human subjects.}

\coi{We declare we don't have conflicts of interest.}

\bibliography{sample}





\end{document}